# Registration-Free Hyperspectral Reconstruction from RGB via a Permutation-Invariant Gram-Matrix Principle

Jiangsan Zhao[1,*], Masayuki Hirafuji[2], Seishi Ninomiya[2,3], Jakob Geipel[1], Wei Guo[2]

[1] *Department of Agricultural Technology, Norwegian Institute of Bioeconomy Research (NIBIO), Ås, Norway*
[2] *Laboratory of Field Phenomics, Graduate School of Agriculture and Life Sciences, The University of Tokyo, Nishitokyo, Tokyo 188-0002, Japan*
[3] *Plant Phenomics Research Center, Nanjing Agricultural University, Nanjing, China*

* Corresponding author. E-mail: jiangsan.zhao@nibio.no. Other e-mails: hirafuji@g.ecc.u-tokyo.ac.jp, snino@g.ecc.u-tokyo.ac.jp, guowei@g.ecc.u-tokyo.ac.jp, jakob.geipel@nibio.no.

***Abstract*—**Reconstructing a spatially and spectrally high-resolution hyperspectral image (HR-HSI) by fusing a low-spatial-resolution hyperspectral image (LR-HSI) with a high-spatial-resolution RGB image (HR-RGB) normally rests on two fragile assumptions: that the two images are precisely registered, and that the camera response function (CRF) relating spectra to RGB is known. Both are hard to guarantee when the modalities come from different sensors, and methods that depend on them degrade sharply when the assumptions break. We remove both assumptions with a single idea: the Gram matrix of an unmixed abundance map is invariant to any permutation of pixels, so it compares two images of the same scene purely through their shared material composition—independent of pixel ordering and registration, and without prior knowledge of the RGB sensor response. Supervising an RGB-to-HSI mapping by matching this permutation-invariant statistic makes the learning objective exactly invariant to pixel ordering and independent of spatial correspondence. The effect is stark: under a full random permutation of the HR-RGB pixels, a state-of-the-art fusion method collapses, with its correlation with the ground truth falling from 0.998 to essentially zero, whereas, after inverse reindexing for evaluation, our reconstruction is unchanged. Building on this principle, a residual spectral super-resolution function maps the HR-RGB directly to the HR-HSI, trained without registration, without a known CRF, and without paired supervision. Across indoor, natural-scene, and remote-sensing benchmarks, it attains accuracy comparable to methods that require both assumptions while remaining robust when those assumptions are violated. An extensive ablation further shows that reconstruction accuracy is largely insensitive to the specific loss used to match the Gram matrix, indicating that performance stems from the permutation-invariant principle itself rather than a tuned objective. Code is released at https://github.com/ZJiangsan/UISSF.

***Index Terms*—**Hyperspectral reconstruction, image fusion, permutation invariance, Gram matrix, registration-free, camera response function, unsupervised learning, spectral super-resolution.

## I. Introduction

Hyperspectral images (HSIs) record the spectral signature of materials across many contiguous narrow bands spanning the visible to short-wave infrared range. This dense spectral sampling carries far more diagnostic information than a conventional three-channel RGB image [1], and has driven the adoption of HSIs across remote sensing, agriculture, ecology, medicine, and industrial inspection [2]–[8].

In practice, however, the instruments that acquire HSIs force a compromise between spatial, spectral, and temporal resolution. Scanning spectrometers reach high spectral fidelity but are slow and cumbersome [9], whereas snapshot devices capture a scene in one exposure at the expense of spatial detail [10], [11]. As a result, an image that is simultaneously high in spatial and spectral resolution (HR-HSI) is difficult to obtain directly. Computational alternatives have therefore been pursued, among which coded-aperture designs [12]–[14] and fusion-based snapshot cameras [11], [14], [15] are prominent, though their elaborate optics and cost limit wide deployment.

HSI super-resolution (HSI-SR) sidesteps these hardware constraints by reconstructing an HR-HSI from two inexpensive and widely available inputs: an LR-HSI and an HR-RGB of the same scene [16]–[21]. Yet almost all such methods carry two assumptions that, crucially, correspond to the two things that are hardest to obtain in a real acquisition. First, the LR-HSI and HR-RGB are assumed to be spatially registered, so that corresponding pixels can be paired during learning. Pixel-accurate registration between a hyperspectral sensor and an RGB camera is notoriously difficult in practice: the two use different optics and detectors, are seldom perfectly co-mounted, often image at different times, and are subject to parallax, so residual misalignment is the norm rather than the exception. Second, the CRF that maps spectral radiance to RGB values is assumed to be known, so that one input can be projected onto the other. Yet CRFs are frequently proprietary, undocumented for consumer cameras, or recoverable only through a laboratory calibration that most end users cannot perform. These are therefore not incidental technicalities but the principal obstacles to deploying HSI-SR outside curated datasets: a method that presumes both requirements may perform well on data where alignment and calibration have been arranged offline, yet degrade sharply—or fail outright—on the uncalibrated, imperfectly registered inputs encountered in the field. The two requirements effectively render such methods supervised or pseudo-unsupervised, and confine them to the very conditions that practical use cannot guarantee.

Existing attempts to relax the registration assumption still treat misalignment as something to be measured and corrected—estimating a rigid or non-rigid warp, or learning a deformable alignment, and aligning the inputs before or during reconstruction [53]. We take a different stance: rather than recovering spatial correspondence, we discard it. The basis for doing so is a supervisory signal that is invariant to where pixels lie. We observe that the Gram matrix of an abundance map—obtained by linearly unmixing an image into a set of endmembers and their per-pixel weights—depends only on how materials co-occur across the scene, not on the spatial arrangement of the pixels. Two images of the same scene therefore yield matching abundance Gram matrices even if one is shifted, rotated, resampled to a different size, or has its pixels randomly shuffled.

This invariance is a property of the operator, not of a particular task, and so it is indifferent to the direction of the spectral mapping. The same abundance-Gram-matrix supervision can drive the well-posed contraction from spectra to RGB—which amounts to recovering the CRF, and which we established in earlier work [23]—and the inverse, far harder expansion from RGB to spectra, which is the genuinely ill-posed problem at the heart of HSI-SR. We treat this as one principle with two directions: the contraction is the easy special case, and this paper uses the same correspondence-free invariance to solve the difficult expansion directly. An autoencoder first unmixes the LR-HSI into endmembers and an abundance map. A residual spectral super-resolution function (SSF), a linear network that maps the HR-RGB to the target HR-HSI while leaving spatial structure untouched, is then optimized so that the abundance map of its output and that of the reference LR-HSI produce the same Gram matrix. The RGB channels of the input serve as a strong starting cue for the residual mapping, which proves essential for stable learning. The optimized network weights constitute the learned SSF, which directly generates the HR-HSI from any given HR-RGB.

This work makes four contributions. **(i)** We identify and formalize the permutation invariance of the abundance Gram matrix as a correspondence-free principle for spectral mapping, and show that one operator supervises the mapping in either direction—the well-posed contraction to RGB and the ill-posed expansion to spectra—thereby removing the registration requirement from direct HSI-SR. **(ii)** We design a residual SSF that performs the harder, low-to-high spectral expansion end-to-end in a fully unsupervised and sensor-independent manner. Here, sensor-independent means that no known, predefined, selected, or estimated RGB camera response function is supplied; the method also requires no downstream reconstruction model. **(iii)** Through extensive experiments, including a pixel-permutation stress test that destroys all spatial correspondence, we show that the learned SSF is invariant to pixel ordering: under a full random permutation of the input pixels, a state-of-the-art fusion method collapses—its correlation with the ground truth falling from 0.998 to essentially zero—while, after inverse reindexing for evaluation, our reconstruction is unchanged. On aligned data the method still reaches accuracy comparable to state-of-the-art methods that depend on registration and known CRFs. Because it requires neither a known CRF nor registered inputs, the method applies directly to uncalibrated, imperfectly registered acquisitions—the regime in which registration- and calibration-dependent methods lose their guarantees—giving it a practical reach that, to our knowledge, prior HSI-SR methods do not offer. **(iv)** An extensive loss ablation shows that reconstruction accuracy is largely insensitive to the specific discrepancy used to match the abundance Gram matrix, indicating that performance stems from the permutation-invariant principle itself rather than from a carefully tuned loss.

Although this work focuses on hyperspectral image super-resolution, the underlying correspondence-free supervision is independent of the specific reconstruction task. Because it compares images through a permutation-invariant representation rather than pixel correspondence, the same principle may be applicable to other multimodal remote-sensing reconstruction problems in which accurate cross-modal registration is unavailable. Exploring such extensions is left for future work.

***Relation to prior work.*** The abundance-Gram-matrix operator and its use for spectral mapping were introduced in our earlier study on CRF learning [23], which addressed the contraction from spectra to RGB and produced a CRF to assist a separate, existing HSI-SR model. We treat that result as the established, well-posed special case of a single principle and do not re-report it here; readers are referred to [23] for its method and results. The present paper makes the principle explicit by formalizing the permutation invariance that underlies it, and applies the same correspondence-free supervision to the inverse and ill-posed expansion from RGB to spectra. This direction is solved end-to-end by a residual SSF that produces the HR-HSI directly, with no downstream model. The shared element across the two directions is precisely the unifying operator; the formulation, the residual network, the ill-posed problem it solves, and the invariance result that ties the two directions together are new.

The remainder of this paper is organized as follows. Section II surveys fusion-based HSI-SR and the registration and CRF assumptions that constrain it. Section III formalizes the problem. Section IV presents the decomposition network, the correspondence-free principle, the residual SSF, and the training objective. Section V reports experiments on three benchmarks. Section VI concludes.

## II. Related Work

We organize this survey around the two assumptions that determine whether an HSI-SR method can be applied to real acquisitions: whether it requires the HR-RGB and LR-HSI to be spatially registered, and whether it requires a known camera response function. We deliberately foreground this dependence axis rather than the more familiar supervised-versus-unsupervised distinction. The latter records only whether a method uses external training targets, a requirement that recent work has largely overcome; it does not capture what actually limits deployment. A method may be fully unsupervised yet still fail on field data because it presumes registration and a known CRF, whereas a supervised method may tolerate misalignment if its supervision is constructed at low resolution. Supervised or not is therefore a poor predictor of practical applicability; registration and CRF dependence is the informative axis, and we classify prior work accordingly, invoking the supervised-unsupervised distinction only where it explains why a particular method exhibits a given dependence.

Fusion-based HSI-SR reconstructs an HR-HSI from a co-scene LR-HSI and HR-RGB, both obtainable at low cost from existing devices [16], [17]. Two broad families exist. The first is statistical: pansharpening-style fusion [24], [25], matrix factorization [26], [27], and tensor representations [28]–[31]. The second is based on deep convolutional networks [16], [32], [33]. Despite their differing machinery, the great majority assume a strict spatial registration between the HR-RGB and the LR-HSI [16], [17], [28]–[34], so that the two inputs can be combined pixel by pixel.

Considerable effort has gone into loosening this requirement, yet it remains anchored in the notion of recovering correspondence. Rigid displacements have been resolved jointly with reconstruction [35], [36], while non-rigid ones are typically handled by a register-then-reconstruct pipeline [37], [38]. In every case the spatial relationship between the inputs must first be determined, and the quality of the result is bounded by the quality of that alignment. Our method departs from this line entirely: it never estimates a warp, because its objective does not depend on spatial correspondence at all.

More recent work has pushed the unregistered setting considerably further while preserving its core premise that misalignment must be estimated and compensated. Deep approaches replace explicit pre-registration with learned alignment—optical-flow predictors, spatial-transformer modules, or deformable aggregation on deep features. The recent UAFL framework [53], the closest method to ours, is representative: it likewise unmixes the inputs and operates in abundance space, but to exploit the unregistered reference it introduces a coarse-to-fine deformable aggregation module that estimates a pixel-level flow and performs sub-pixel refinement. Its design therefore presumes that misalignment must be compensated before fusion. A parallel line addresses the second assumption, the unknown sensor response, through blind or test-time-adaptive fusion that copes with unknown degradation [54], and recent analysis has begun to establish recoverability guarantees for the unregistered fusion problem itself [55]. Our work is complementary to, and conceptually distinct from, these lines. Rather than estimating and correcting misalignment as [53] does, we remove the alignment step altogether by supervising through a permutation-invariant quantity; and rather than treating the unknown CRF as a degradation to be modelled, we make learning independent of it. The distinction is structural rather than merely empirical: any method that recovers correspondence presupposes that a correspondence exists to be found, whereas a full pixel permutation—the worst case we evaluate in Section V-D—leaves no correspondence to recover, yet leaves the abundance Gram matrix unchanged. The honest trade-off is that alignment-based methods may retain higher peak fidelity under accurate registration, while our formulation is uniquely invariant to arbitrary misalignment.

A second and largely independent limitation is the reliance on a known CRF to relate spectral radiance to RGB values [11], [17]–[19], [39]–[44]. Reconstruction accuracy is strongly sensitive to the CRF used to form the RGB input [16], [23], [45], [46], yet the CRF of a given set of RGB images is usually unavailable once the source camera is unknown. Dictionary- and selection-based remedies presuppose access to a calibrated CRF dictionary [16], which is itself a strong assumption. The method proposed here removes this dependence: no known, predefined, selected, or estimated CRF enters the learning process. By making supervision blind to spatial layout and free of any CRF, we relax both of the assumptions that have constrained the field, and we validate the approach against two representative methods with different dependence profiles, u2MDN and WLRTR.

## III. Problem Formulation

Let the HR-RGB be $X \in \mathbb{R}^{HW\times l}$, where $H$, $W$ are its height and width and $l = 3$ is the number of colour channels. Let the LR-HSI be $Y \in \mathbb{R}^{hw\times L}$, with height $h$, width $w$, and $L = 31$ spectral bands. The goal is to learn a spectral super-resolution function that maps the HR-RGB to its hyperspectral counterpart, the target HR-HSI of size ${}^{HW\times L}$.

An HSI can be expressed by linear unmixing as a small set of endmembers—characteristic spectral signatures of the materials present—together with an abundance map giving the per-pixel weight of each endmember [20]. For the LR-HSI,

$$Y(hw\times L) = A(hw\times\theta) \cdot E(\theta\times L), \qquad (1)$$

where $A(hw\times\theta)$ is the abundance map, $E(\theta\times L)$ is the endmember matrix, and $\theta$ is the number of endmembers, chosen according to scene complexity. Because the HR-RGB and the LR-HSI depict the same scene, the same materials are present in both. Rather than assuming this, we enforce it: the endmember matrix E is estimated once from the LR-HSI and then held fixed, so that the target HR-HSI is constrained to share exactly those endmembers and differs only in its abundances. The target HR-HSI therefore decomposes with the same endmembers,

$$Y(HW\times L) = A(HW\times\theta) \cdot E(\theta\times L), \qquad (2)$$

where $A(HW\times\theta)$ is the abundance map of the target HR-HSI and $\theta$ matches the endmember count of the LR-HSI. The two abundance maps differ in the number of rows (pixels) but share the column dimension $\theta$, a property exploited below to compare them without pixel correspondence.

## IV. Proposed Approach

### A. Overview

The model comprises two subnetworks (Fig. 1): an autoencoder that unmixes the LR-HSI into endmembers and an abundance map, and a residual SSF that maps the HR-RGB directly to the target HR-HSI. The autoencoder is trained first. Its encoder is then reused, with frozen weights, to decompose the SSF output, so that the abundance maps of the generated HR-HSI and of the reference LR-HSI are produced by the same operator. The SSF is optimized until the Gram matrices of these two abundance maps agree.

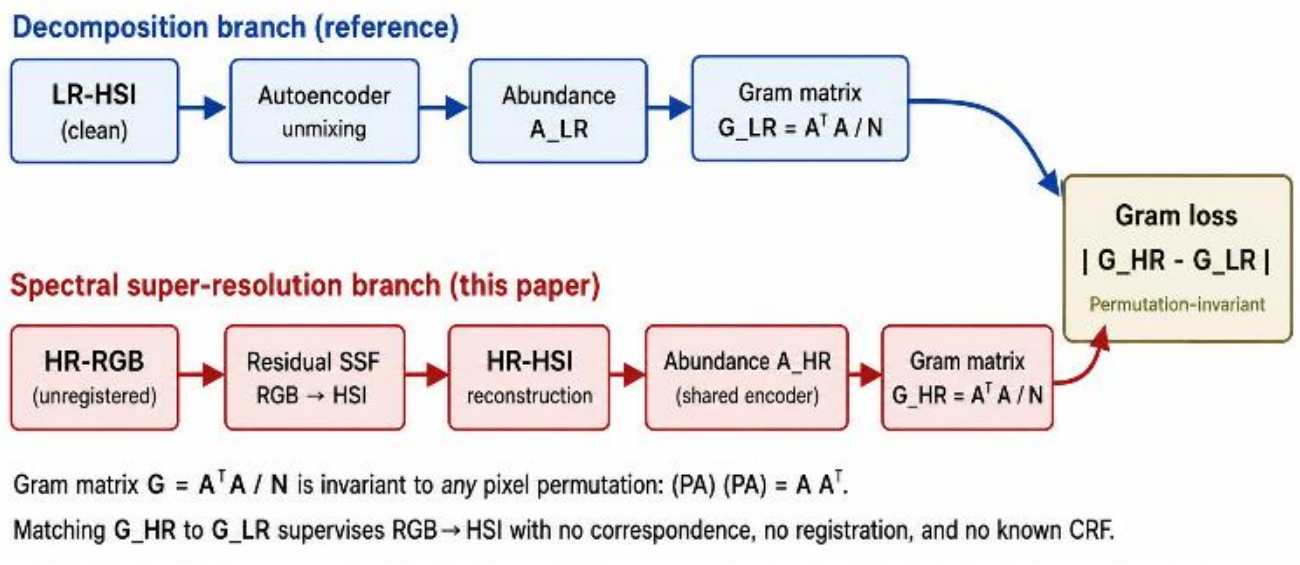

**Fig. 1.** Flowchart of the UISSF residual learning network.

### *B. LR-HSI Decomposition*

The decomposition autoencoder is shown in Fig. 2. The encoder is a densely connected linear network that maps each input spectrum to a latent representation; dense connections between layers strengthen the final representation. A stick-breaking transform then converts this representation into a valid abundance vector that is non-negative and sums to one. The number of endmembers is set to $\theta = 40$, chosen to exceed the 31 spectral bands of the target so that the endmember basis is expressive enough to span the data. The decoder is a single linear layer of 31 nodes that reconstructs the input spectrum from the abundance vector; its 40×31 weight matrix is precisely the endmember set, with 40 endmembers each defined over 31 bands.

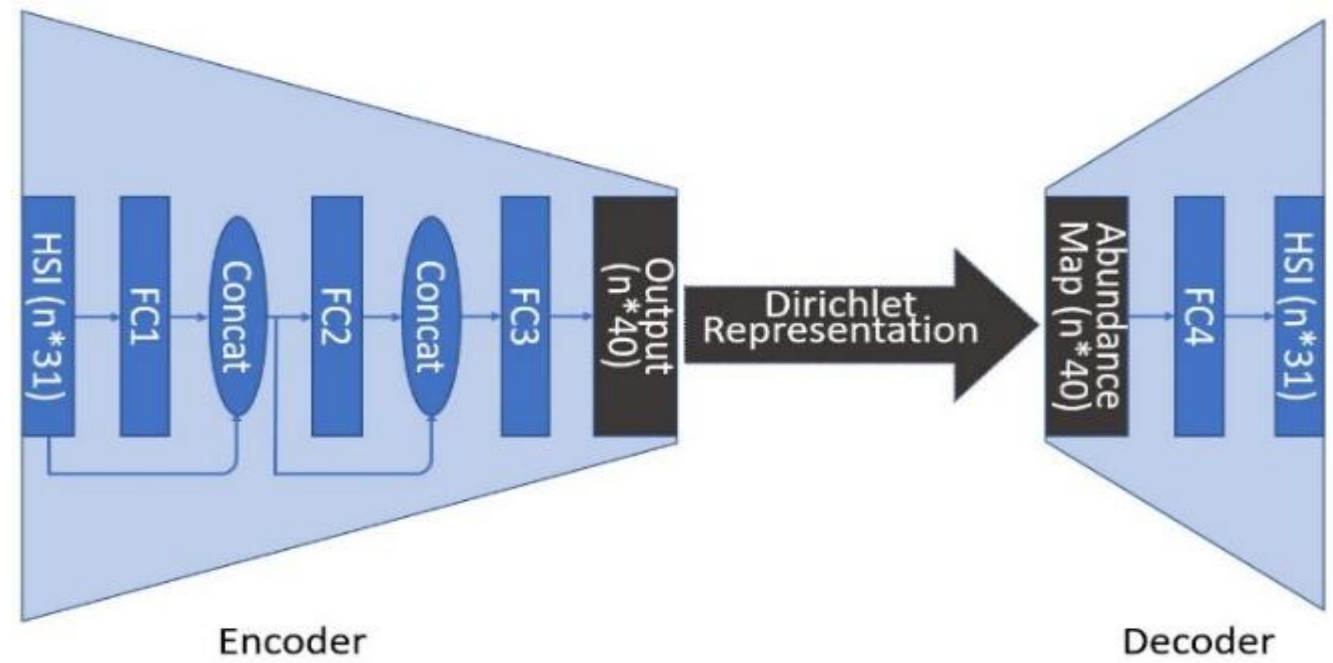

**Fig. 2.** Structure of the encoder-decoder used for LR-HSI decomposition.

### *C. Correspondence-Free Supervision via the Gram Matrix*

The two abundance maps to be compared, $A(HW\times\theta)$ for the target HR-HSI and $A(hw\times\theta)$ for the reference LR-HSI, have different numbers of rows because the images differ in spatial resolution, and, crucially, their rows are not in correspondence because the inputs are unregistered. Standard pixelwise losses therefore cannot be applied. We resolve both obstacles with a single device. For an abundance map A with N rows and $\theta$ columns, define its (normalized) Gram matrix as

$$G = (1/N) \cdot A^T A \in \mathbb{R}(\theta\times\theta). \tag{3}$$

Two properties follow immediately. First, G has size $\theta\times\theta$ regardless of N, so the Gram matrices of the HR and LR abundance maps share identical dimensions and can be compared by any ordinary loss. Second, and more importantly, G is invariant to any permutation of the rows of A. For any permutation matrix P,

$$(PA)^T (PA) = A^T P^T P A = A^T A, \tag{4}$$

since $P^T P = I$. Equation (4) holds for any permutation matrix P and therefore certifies exact invariance to a complete random shuffle of the image. Full random permutation removes all pixelwise correspondence, local neighbourhoods, edges, shapes, and spatial structure while preserving the complete set of observed pixels. Unlike ordinary geometric misalignments, which generally retain substantial spatial organization, it therefore provides a particularly stringent test of whether the learning objective depends on correspondence. The abundance Gram matrix characterizes a scene through the co-occurrence statistics of its materials, independent of spatial layout. This is why registration becomes unnecessary: matching abundance Gram matrices constrains the mapping to reproduce the correct spectral composition of the scene without requiring the HR-RGB and LR-HSI to be spatially aligned. We verify this empirically in Section V-D by learning the SSF from a fully pixel-permuted HR-RGB (Table V), the experimental counterpart of Eq. (4). A third property is equally consequential: the operator constrains the shared composition of two co-scene images and is indifferent to which one is the source and which the target, so the same supervision applies whether the mapping contracts spectra to RGB or expands RGB to spectra. The contraction is the well-posed case treated in [23]; below we use the identical operator to drive the ill-posed expansion. We further exploit the coupling of spatial and abundance information that the Gram matrix captures, which has proven effective for representing such maps [47].

### *D. Residual SSF Learning*

Because the endmembers are shared by construction, the encoder and decoder learned in the decomposition stage are frozen during SSF learning, fixing the endmember basis; only the abundances of the generated HR-HSI are free to adapt. The shared-endmember property is therefore enforced rather than assumed, and the close agreement between the reconstructed and ground-truth spectra (Figs. 5–7) confirms that constraining composition in this way recovers the correct spectral content. The SSF is realized as a single-layer linear residual network that maps the three-channel HR-RGB to the 31-channel target HR-HSI without altering spatial structure (Fig. 3). The residual formulation lets the original RGB channels serve as a starting cue, which markedly eases optimization relative to learning the mapping from scratch. The network is trained by passing its output through the frozen encoder and minimizing the discrepancy between the resulting abundance Gram matrix and that of the LR-HSI; at convergence, the network weights are the learned SSF, applicable to any HR-RGB of the scene.

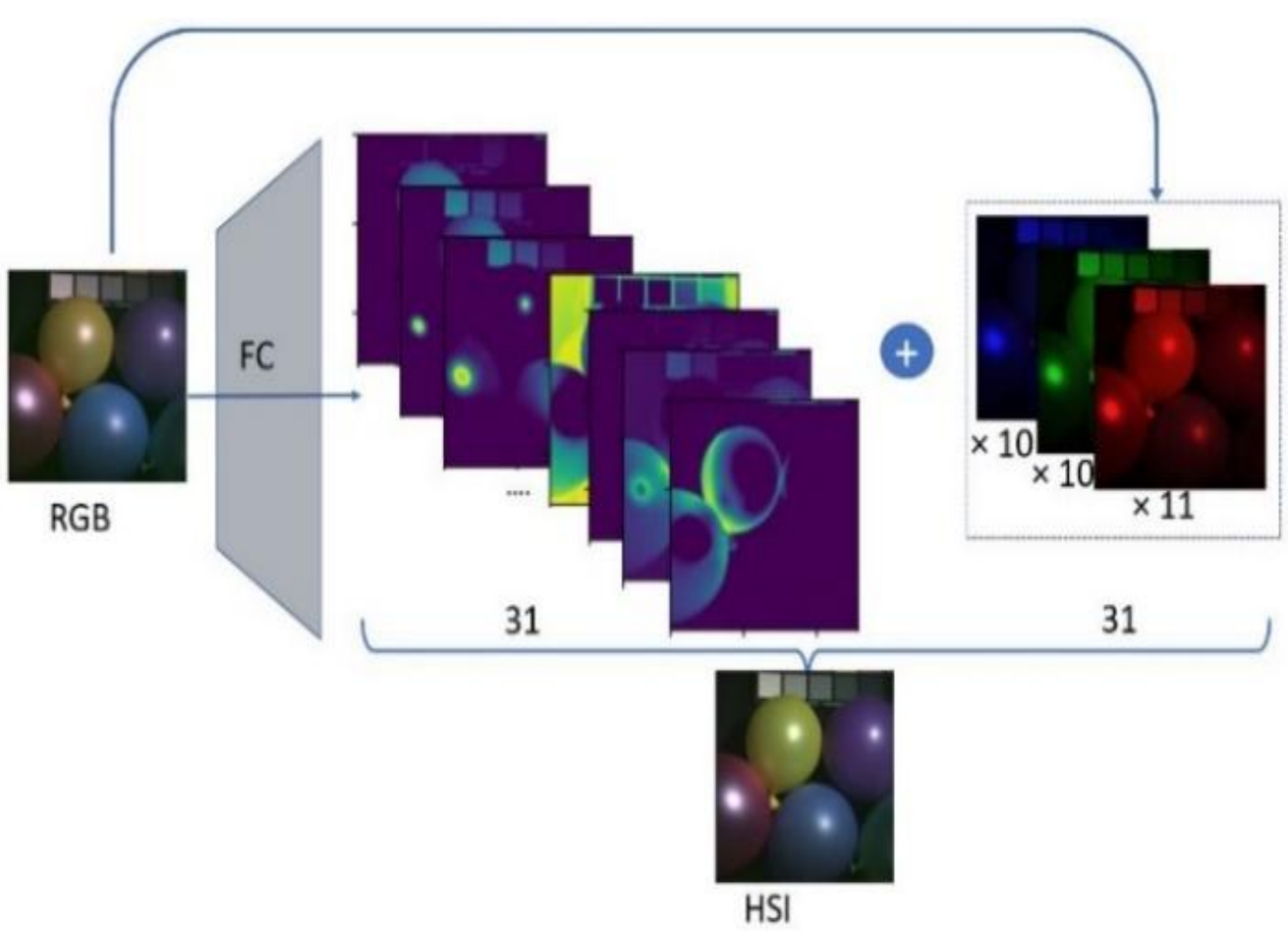

**Fig. 3.** Structure of the residual SSF.

### *E. Inference: Generating the HR-HSI*

At inference the trained components are composed into a single feed-forward pass that maps an HR-RGB to its HR-HSI. The HR-RGB is first passed through the learned residual SSF, yielding an initial spectral estimate; this estimate is then projected onto the endmember manifold recovered in the decomposition stage by encoding it into an abundance map with the frozen encoder and reconstructing the spectrum with the frozen decoder. Concretely, the reconstruction is HR-HSI = D(E(f(X))), where X is the HR-RGB, f is the residual SSF, and E and D are the encoder and decoder trained solely on the LR-HSI in the decomposition stage. Because E and D are optimized only on the LR-HSI, an input that is always available, this projection introduces no dependence on the ground-truth HR-HSI and is valid at deployment. Passing the SSF output through the shared decomposition in this way constrains every reconstructed pixel to lie in the span of the scene endmembers, which suppresses residual spectral error and yields the reported reconstructions. All quantitative results in Section V are computed from this decoder-path output.

### *F. Loss Function*

The low-to-high spectral expansion is severely ill-posed and benefits from a supervisory signal that constrains the Gram-matrix match from complementary angles. We use a composite loss combining a collaborative ℓ2,1-norm term, an ERGAS term, and a SAM term, all evaluated between the two abundance Gram matrices. The ℓ2,1-norm applies an ℓ2-norm across each vector followed by an ℓ1-norm, regularizing magnitude during spectral recovery [17]. ERGAS measures global relative error [48], and is well suited to capturing small differences when absolute magnitudes are low. SAM measures the angular difference in vector shape [52] and is complementary to the magnitude-oriented ℓ2,1 and ERGAS terms. Combining the three imposes stricter, mutually reinforcing constraints, steering the optimization toward the correct solution. Let ΔG = GHR − GLR. The terms are given in (5)–(7) and the total loss in (8).

$$\mathscr{L}_{2,1} = \sum_i \left(\sum_j \Delta G_{ij}^2\right)^{1/2} \quad (5)$$

$$\mathscr{L}ERGAS = (100/\gamma)\left[(1/m)\sum_i\sum_j\left(\Delta G_{ij}/(G^{LR}{}_{ij}+\varepsilon)\right)^2\right]^{1/2} \quad (6)$$

$$\mathscr{L}SAM = (1/m)\sum_i \arccos\left(\langle g_i^{HR}, g_i^{LR}\rangle/(\|g_i^{HR}\|_2\|g_i^{LR}\|_2+\varepsilon)\right) \quad (7)$$

where GHR and GLR are the abundance Gram matrices of the target HR-HSI and reference LR-HSI, respectively; $g_i$ denotes the ith row of a Gram matrix; m is the number of rows; γ is the super-resolution scale factor (8 in all experiments); and ε is a small positive constant for numerical stability. The total loss sums the three components,

$$Loss_total = \lambda_1 \cdot \ell 2,1\ loss + \lambda_2 \cdot ERGAS\ loss + \lambda_3 \cdot SAM\ loss, \quad (8)$$

with weights $\lambda_1$, $\lambda_2$, $\lambda_3$ balancing the components.

### *G. Training*

Input spectra of the LR-HSI and HR-RGB were scaled to the 0–1 range and zero-centred before entering the networks. The encoder width was fixed at 40 output nodes after preliminary screening. The decoder and SSF weights were initialized using the PyTorch defaults [50]. The relatively easy decomposition autoencoder was trained with a combination of ℓ2,1-norm and SID losses, which prior work found effective, while the SSF was trained with the composite ℓ2,1–ERGAS–SAM loss above. The component weights $\lambda_1$, $\lambda_2$, and $\lambda_3$ were each set to one; tuning them could further improve results but lies beyond the present scope. Both stages were run until the loss showed no further decrease over 1000 epochs.

## V. Experiments and Results

### *A. Datasets and Experimental Setting*

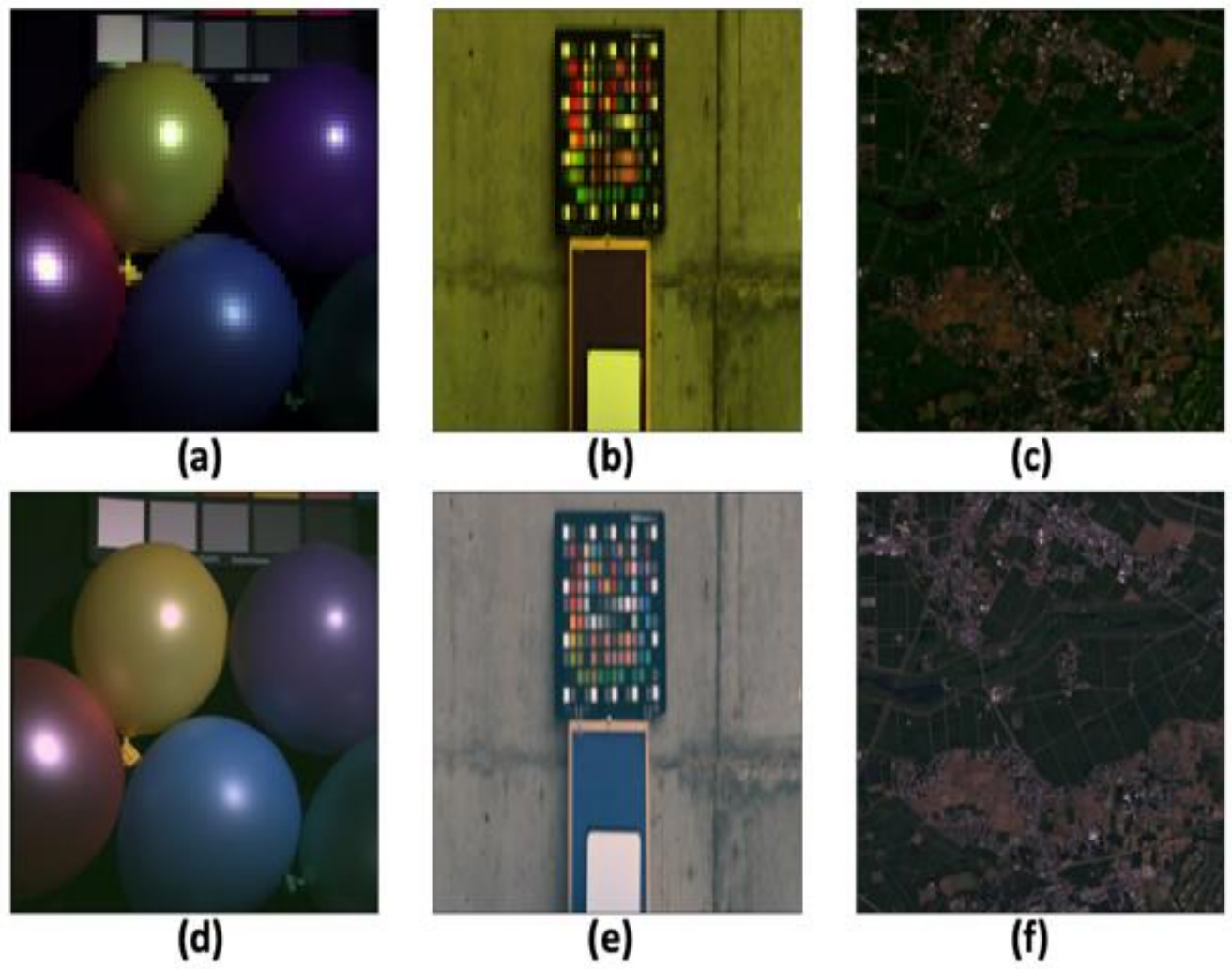

**Fig. 4.** Example datasets used for visualization. (a–c) true-colour RGB of the LR-HSI of CAVE, ICVL, and Chikusei; (d–f) natural-colour HR-RGB of CAVE, ICVL, and Chikusei.

Three widely used benchmarks were evaluated, spanning indoor, natural-scene, and remote-sensing imagery: CAVE [51], ICVL [18], and Chikusei [20]. CAVE contains 32 indoor scenes of 512×512 pixels with 31 bands from 400 to 700 nm. ICVL comprises natural scenes of 1000×1000 pixels with 31 bands. Chikusei is an airborne scene of 2517×2335 pixels and 128 bands over 363–1018 nm; a central 1000×1000 patch was cropped and the 31 bands within 400–700 nm were used. To form RGB inputs and to probe CRF dependence, the standard CRFs of the Nikon 700D [17] and Canon 500D [44] cameras were applied. LR-HSIs were obtained from the HR-HSIs with a Gaussian filter whose width matched the super-resolution scale and whose standard deviation was 0.5; a scale of 8 was used throughout. To assess robustness to imaging noise, zero-mean Gaussian noise was added to the [0,1]-scaled HR-RGB at signal-to-noise ratios of 40, 30, and 20 dB before the inputs were zero-centred, with the noise specified by $\mathrm{SNR} = 10 \cdot \log_{10}(\text{mean signal power} / \text{noise variance})$; the LR-HSI was left noise-free, as it originates from a controlled spectrometer rather than a consumer camera.

### B. Compared Methods

The comparison methods were chosen for their differing dependence on the two assumptions (Table I). WLRTR [29] requires both registration and a known CRF; u2MDN [17] requires a known CRF but tolerates unregistered inputs because it is supervised on constructed low-resolution pairs; EDIP-Net [56] is unsupervised yet still requires registration and estimates the CRF; and the proposed UISSF requires neither. This spread is deliberate: it shows that the registration and CRF dependence persists across statistical, supervised, and unsupervised paradigms alike, and that UISSF removes both regardless of paradigm. The baselines were run with their recommended hyperparameters and tuned for best performance. To probe robustness to misalignment, the HR-RGB was additionally rotated by 90° counter-clockwise before reconstruction. A fair quantitative comparison requires running each competing method under identical conditions, which in turn requires a public implementation; we therefore benchmark against methods whose code is available. The most recent unregistered methods, including UAFL [53], are positioned conceptually in Section II rather than included in the tables, as their implementations were not publicly available at the time of writing.

**TABLE I**
*Dependence of each method on registration and CRF*

| Model | Registration | CRF |
|---|---|---|
| WLRTR | yes | yes |
| u2MDN | no | yes |
| EDIP-Net | yes | yes (estimated) |
| UISSF | no | no |

### C. Evaluation Metrics

Three complementary metrics quantified reconstruction quality: peak signal-to-noise ratio (PSNR), spectral angle mapper (SAM) [52], and ERGAS [48]. SAM measures the per-pixel spectral angle between prediction and ground truth, while PSNR and ERGAS are MSE-based band-wise indices of spatial fidelity and global quality. Higher PSNR and lower SAM and ERGAS indicate better performance.

### D. Comparison with State-of-the-Art Methods

The decomposition autoencoder reconstructed the LR-HSIs almost perfectly, with low SAM and ERGAS and high PSNR across all three datasets (Table II), confirming that the learned endmembers and abundances faithfully represent the data. WLRTR suffered large degradations whenever registration or CRF was incorrect (Table III), and u2MDN degraded sharply when the CRF was wrong (Table IV); u2MDN showed no registration sensitivity because it is trained in a supervised manner from the LR-HSI and a CRF-converted LR-RGB. In contrast, UISSF is independent of both factors—neither registration nor CRF enters any part of its learning—and its accuracy was comparable to the best results of the two baselines. The relevant question under deliberate misalignment is whether a method remains accurate after all spatial correspondence has been removed. To establish this in the strongest possible form, a fixed random permutation P was applied to all HR-RGB pixels before SSF learning, thereby eliminating every spatial correspondence between the HR-RGB and LR-HSI during optimization. The reconstructed spectra retained the same permuted ordering because the SSF operates independently on each pixel. For visualization and pixelwise metric calculation only, the inverse permutation $P^{-1}$ was then applied to the reconstructed HR-HSI, restoring each predicted spectrum to the original position of its corresponding RGB pixel. Neither the inverse permutation nor the ground-truth HR-HSI was used during learning. Full random permutation destroys all pixelwise correspondence, local neighbourhoods, edges, shapes, and spatial structure while preserving the complete set of observed RGB pixels, making it a stricter stress test of correspondence dependence than ordinary geometric misalignment. Table V is therefore the empirical counterpart of Eq. (4): after inverse reindexing, the reconstruction is unchanged. To make the contrast concrete, we applied the identical fixed permutation to the strongest correspondence-dependent baseline, EDIP-Net, re-ran its full pipeline from scratch on the permuted HR-RGB, and used the same inverse reindexing before evaluation. Although its internal fitting of the unpermuted LR-HSI proceeded normally, its HR-HSI reconstruction collapsed: the correlation coefficient with the ground truth fell from 0.998 on aligned data to essentially zero under permutation, with the spectral angle rising by an order of magnitude. The method continued to optimize its objective and returned a confident output, yet that output was uncorrelated with the true scene because its degradation model presumes a spatial correspondence that the permutation removes. UISSF, by contrast, returned an identical reconstruction to the aligned case after inverse reindexing (Table V), the exact empirical counterpart of Eq. (4). No method dominated on all datasets (Tables III–V), and full visual comparisons appear in Figs. 5–7.

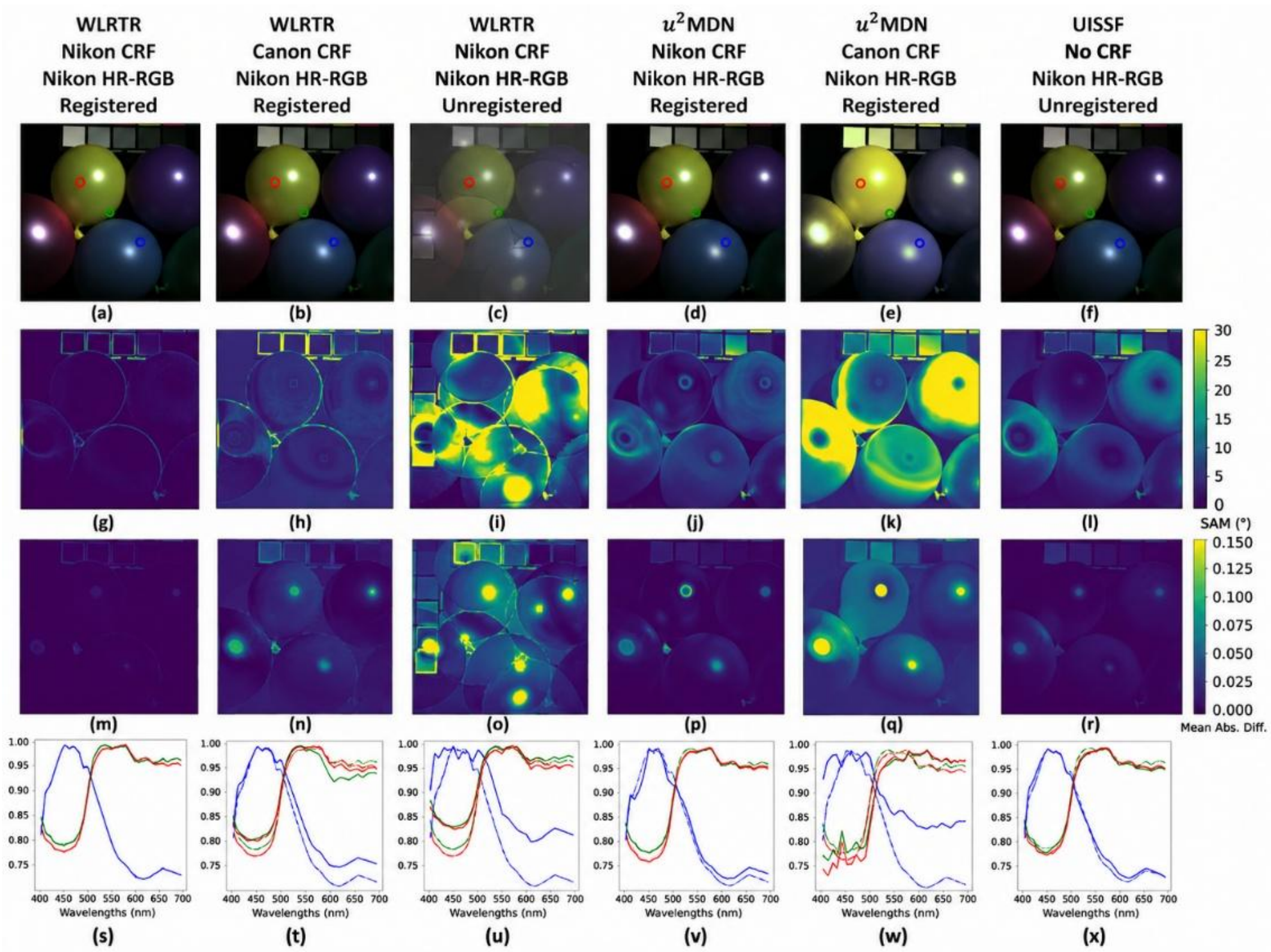

**Fig. 5.** Generated HR-HSI of CAVE by WLRTR, u2MDN, and UISSF. Rows: (a–f) generated HR-HSI; (g–l) SAM versus ground truth; (m–r) mean absolute difference; (s–x) reflectance of three marked pixels. Columns 1–3 are WLRTR (Nikon CRF registered, Canon CRF registered, Nikon CRF unregistered); columns 4–5 are u2MDN (Nikon and Canon CRF, registered); column 6 is UISSF (Nikon HR-RGB, unregistered).

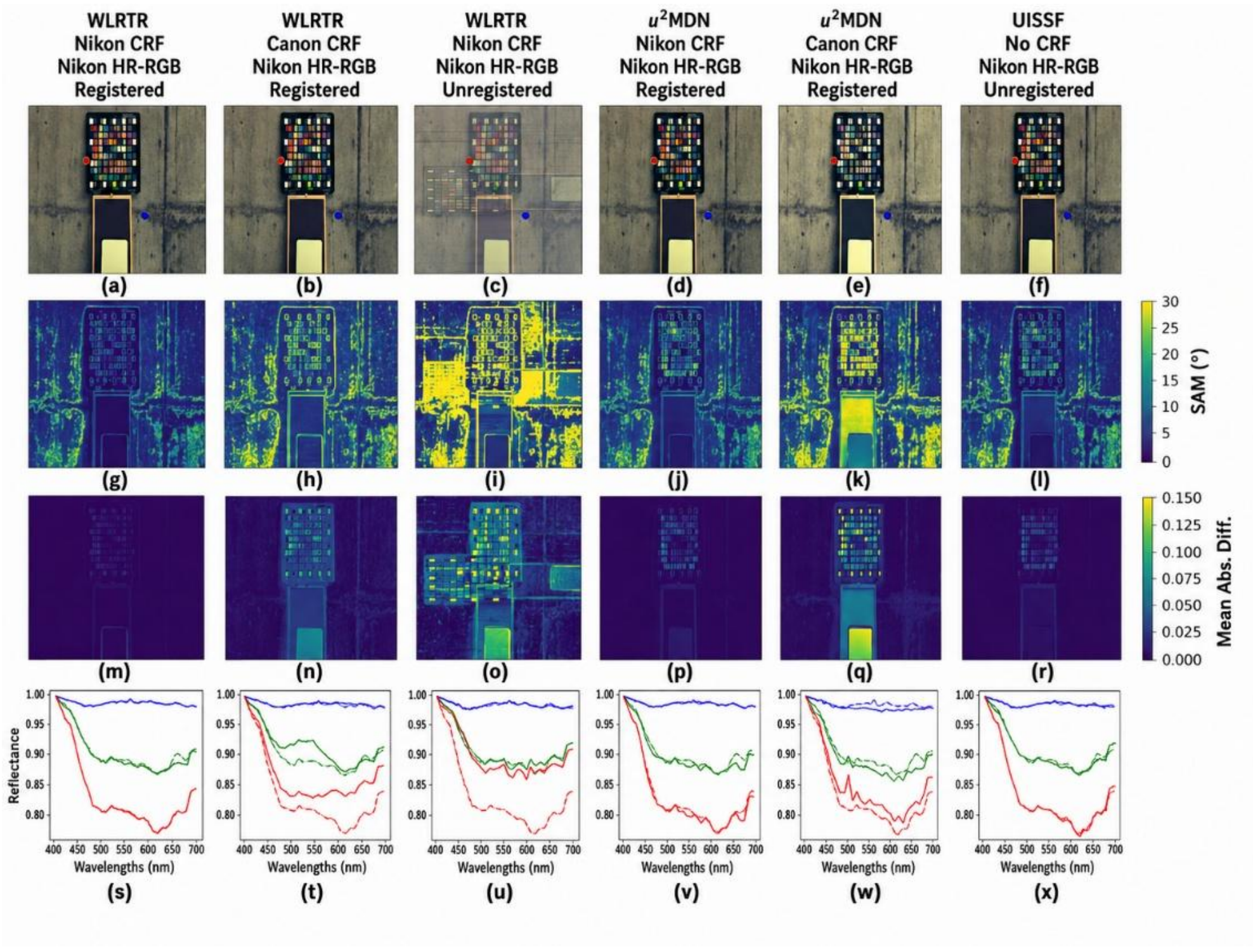

**Fig. 6.** Generated HR-HSI of ICVL by WLRTR, u2MDN, and UISSF, with panels and columns as in Fig. 5.

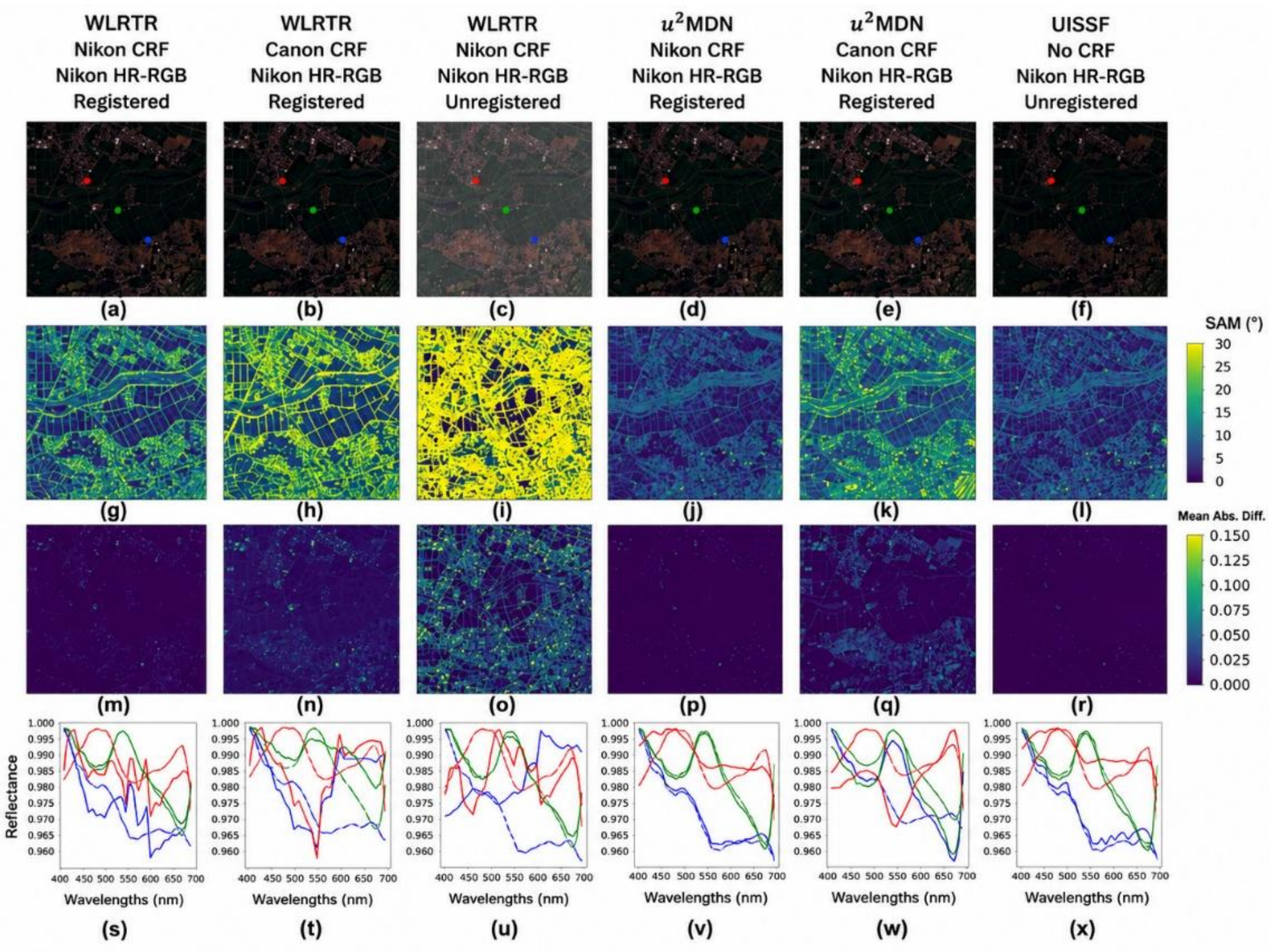

**Fig. 7.** Generated HR-HSI of Chikusei by WLRTR, u2MDN, and UISSF, with panels and columns as in Fig. 5.

**TABLE II**

*Decomposition fidelity of the encoder-decoder on the LR-HSI*

| Dataset | SAM ↓ | PSNR ↑ | ERGAS ↓ |
|---|---|---|---|
| CAVE | 1.04 | 56.58 | 4.80 |
| ICVL | 0.14 | 61.76 | 0.41 |
| Chikusei | 0.37 | 68.20 | 1.88 |

**TABLE III**

*WLRTR with Nikon/Canon CRF on registered and unregistered Nikon HR-RGB*

| CRF / Registration | Dataset | SAM↓ | PSNR↑ | ERGAS↓ |
|---|---|---|---|---|
| Nikon / Reg. | CAVE | 1.26 | 52.57 | 0.96 |
| | ICVL | 5.02 | 46.15 | 1.23 |
| | Chikusei | 8.39 | 45.72 | 5.30 |
| Nikon / Unreg | CAVE | 17.90 | 27.92 | 12.75 |
| | ICVL | 28.67 | 28.29 | 6.54 |
| | Chikusei | 47.16 | 33.27 | 14.59 |
| Canon / Reg. | CAVE | 4.97 | 34.72 | 7.28 |
| | ICVL | 10.58 | 33.87 | 3.83 |
| | Chikusei | 18.18 | 36.65 | 7.22 |

*reg = registered, unreg = unregistered.*

**TABLE IV**

*u2MDN trained with Nikon or Canon CRF, tested on Nikon HR-RGB*

| Dataset | SAM↓ (Nikon) | PSNR↑ (Nikon) | ERGAS↓ (Nikon) | SAM↓ (Canon) | PSNR↑ (Canon) | ERGAS↓ (Canon) |
|---|---|---|---|---|---|---|
| CAVE | 4.92 | 41.12 | 1.29 | 19.35 | 26.50 | 8.23 |
| ICVL | 6.81 | 41.79 | 0.60 | 14.50 | 33.84 | 4.27 |
| Chikusei | 4.86 | 53.15 | 3.59 | 12.25 | 39.23 | 3.79 |

**TABLE V**

*UISSF on pixel-permuted Nikon HR-RGB after inverse reindexing*

| Dataset | SAM ↓ | PSNR ↑ | ERGAS ↓ |
|---|---|---|---|
| CAVE | 4.76 | 43.52 | 6.92 |
| ICVL | 0.80 | 48.63 | 2.76 |
| Chikusei | 2.14 | 53.50 | 6.68 |

### *E. Comparison with a Recent Unsupervised Method on Aligned Data*

To place UISSF against the current state of the art under conditions favorable to competing methods, we compared it with EDIP-Net [56], a recent unsupervised fusion method that reaches high accuracy but, like most current approaches, still estimates the CRF and requires spatially registered inputs. Both methods were evaluated on the same ground-truth HR-HSI with identical degradation (same SRF, same spatial downsampling), so the comparison is on aligned, calibrated data—precisely the regime EDIP-Net is designed for. Table VI reports the result. In this favorable setting the two methods are close: EDIP-Net attains slightly higher peak fidelity, consistent with a multi-stage pipeline that exploits both the estimated CRF and the assumed registration, while UISSF reaches comparable accuracy using neither. The value of UISSF is therefore not a lower error under ideal alignment, but competitive accuracy without the two assumptions EDIP-Net depends on. When spatial correspondence is destroyed, EDIP-Net's degradation model is invalid and its reconstruction collapses, whereas UISSF remains unchanged after inverse reindexing (Table V).

**TABLE VI**

*UISSF vs EDIP-Net [56] on aligned data (same GT, same SRF and downsampling)*

| Dataset | SAM↓ (EDIP-Net) | PSNR↑ (EDIP-Net) | ERGAS↓ (EDIP-Net) | SAM↓ (UISSF) | PSNR↑ (UISSF) | ERGAS↓ (UISSF) |
|---|---|---|---|---|---|---|
| Chikusei | 1.42 | 52.90 | 0.53 | 2.32 | 52.82 | 6.79 |

*EDIP-Net requires an estimated CRF and registered inputs; UISSF requires neither. Both evaluated on identical inputs.*

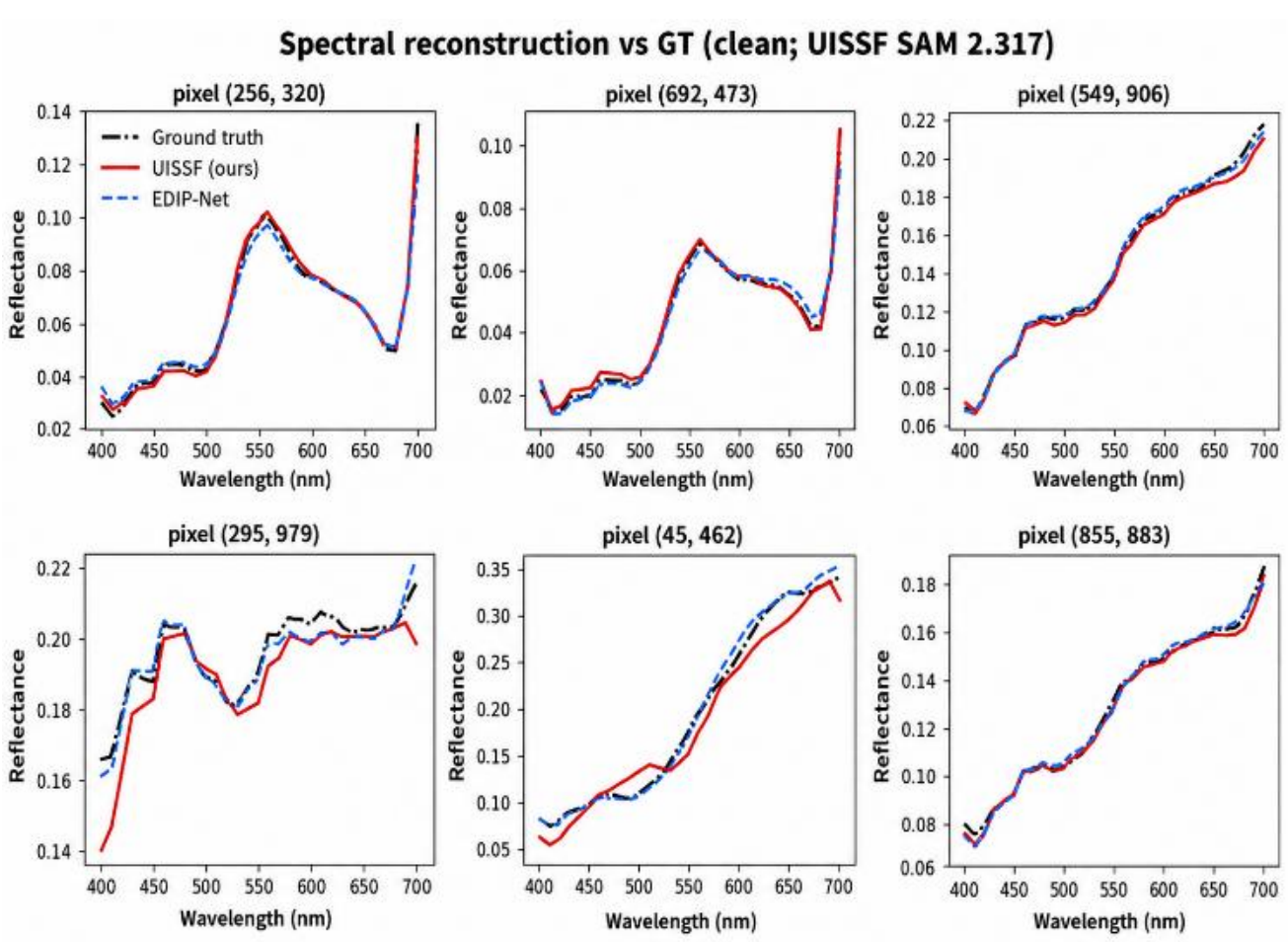

**Fig. 8.** Spectral reconstruction against ground truth on aligned Chikusei at six representative pixels. Ground truth (black dash-dot), UISSF (red), and EDIP-Net (blue dashed) are shown. Both methods track the reference closely; UISSF deviates slightly more on spectrally atypical pixels, consistent with the small difference in mean SAM (2.32 vs 1.42).

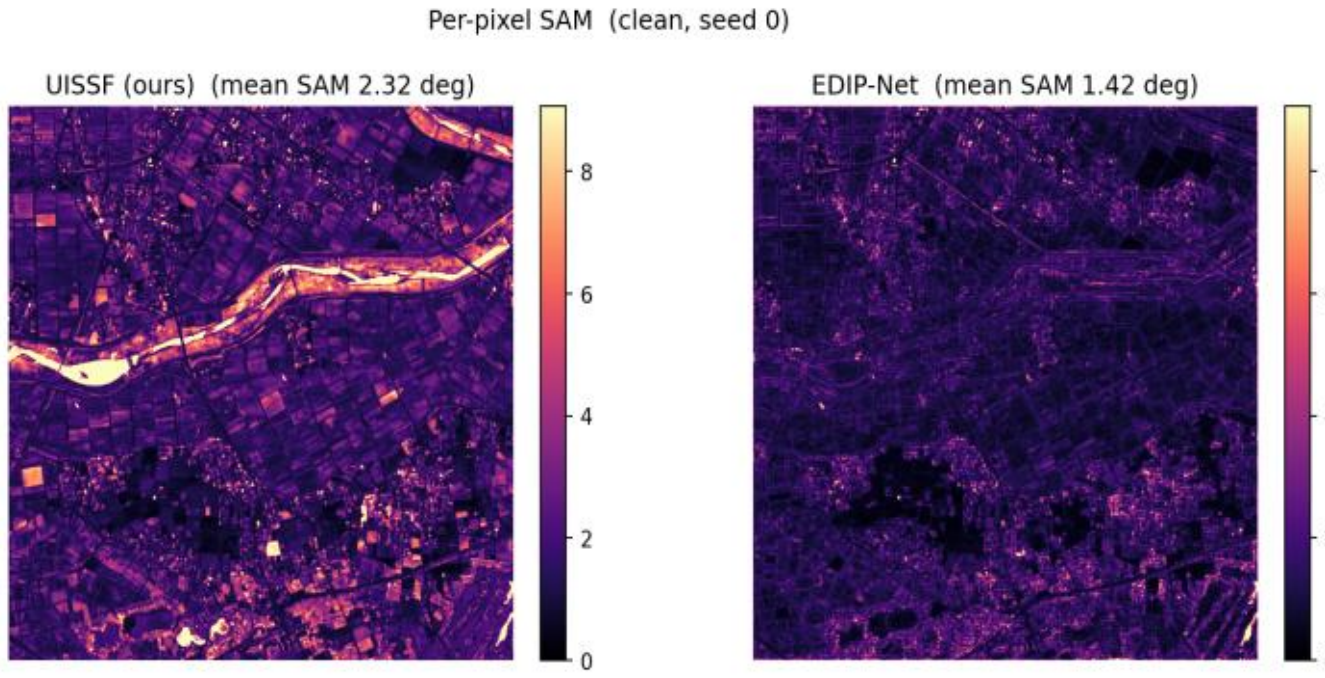

**Fig. 9.** Per-pixel SAM on aligned Chikusei (clean), UISSF (left) and EDIP-Net (right) on a shared colour scale. Both methods reconstruct land cover accurately; UISSF's residual error concentrates on the low-reflectance water bodies, where the spectral angle is intrinsically ill-conditioned and the shared endmember representation is least expressive. This localised difference accounts for most of the gap in mean SAM.

### *F. Ablation on Loss Components*

We adopted a composite $\ell_{2,1}$–ERGAS–SAM loss for the mapping stage, motivated by our experience with the fully supervised LR-HSI decomposition, where combining complementary losses reduced reconstruction error. The ablation in Table VII examines whether this composition is necessary. It is not. Every configuration reaches comparable accuracy: the individual terms, their pairwise combinations, and the full composite all fall within a narrow band (2.13–2.18 SAM, 53.2–53.8 PSNR on Chikusei), and no configuration is meaningfully better than any other. We attribute this to the nature of the supervision. Unlike decomposition, where each loss directly constrains reflectance values, the mapping is supervised through the permutation-invariant abundance Gram matrix; once that statistic is matched, the reconstruction is largely determined, and the particular discrepancy used to match it matters little. The magnitude of the output is fixed not by the loss but by the architecture—the sum-to-one abundance constraint and the physically scaled endmembers of the frozen decomposition—so the loss need only constrain spectral structure, which any of the shape- or relative-error terms accomplishes. This insensitivity is a strength rather than a limitation: it shows that the method's accuracy derives from the permutation-invariant principle itself, not from a carefully weighted objective, and that reconstruction quality is robust to the exact loss composition. We retain the composite loss in all reported experiments for consistency.

**TABLE VII**

*Ablation of loss components on the Chikusei dataset*

| Loss components | SAM ↓ | PSNR ↑ | ERGAS ↓ |
|---|---|---|---|
| $\ell_{2,1}$ | 2.16 | 53.46 | 6.84 |
| ERGAS | 2.18 | 53.27 | 7.00 |
| SAM | 2.14 | 53.24 | 6.70 |
| $\ell_{2,1}$ + ERGAS | 2.18 | 53.32 | 6.95 |
| $\ell_{2,1}$ + SAM | 2.13 | 53.76 | 6.51 |
| ERGAS + SAM | 2.15 | 53.45 | 6.77 |
| $\ell_{2,1}$ + ERGAS + SAM | 2.14 | 53.50 | 6.68 |

### *G. Ablation on the Residual SSF*

Removing the residual connection caused the direct RGB-to-HSI mapping to fail outright (Fig. 10): without the original RGB values as a starting cue, the network could not learn an effective SSF. This confirms that the residual formulation, which lets the input colours anchor the spectral expansion, is essential to the method.

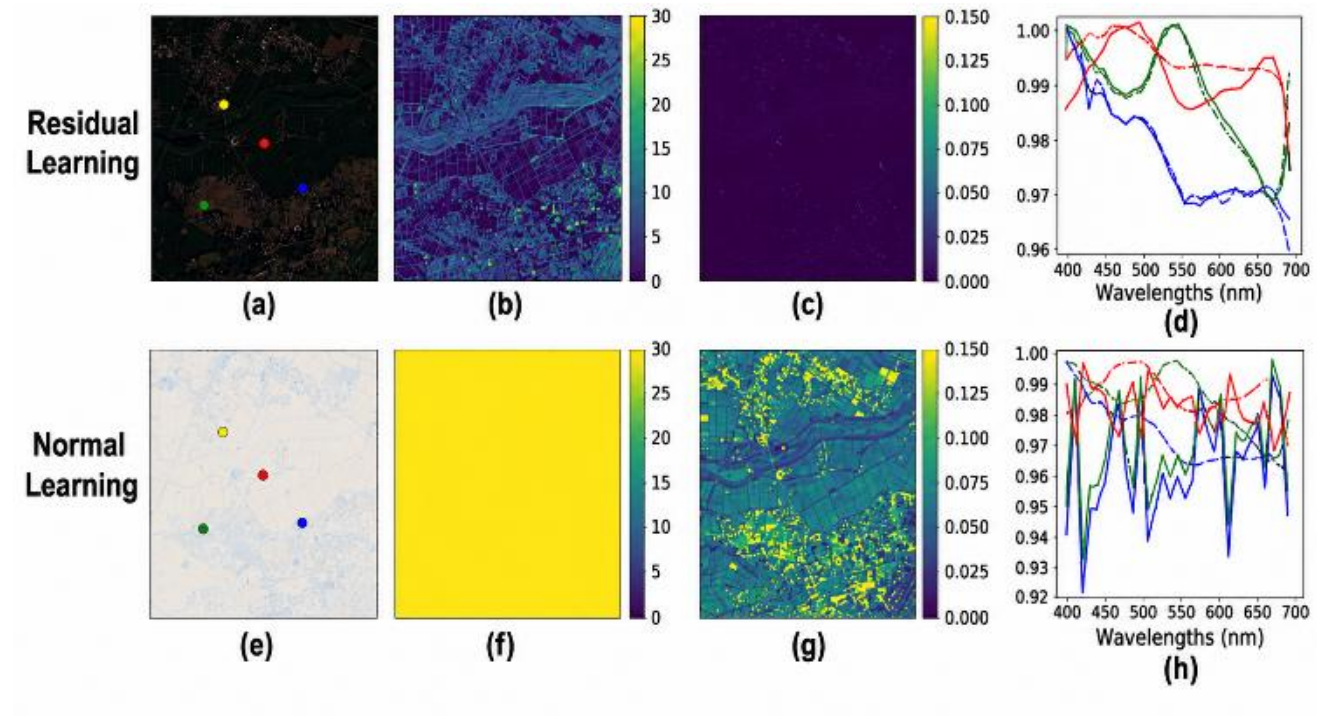

**Fig. 10.** Ablation on the residual SSF using Chikusei. Top row: residual model; bottom row: non-residual model. Columns: generated HR-HSI, SAM, mean absolute difference, and reflectance of three marked pixels.

### *H. Robustness to HR-RGB Noise*

Because the HR-RGB is the input most likely to come from a low-cost consumer sensor, we examined how the learned SSF behaves when that input is noisy. Zero-mean Gaussian noise was added to the [0,1]-scaled HR-RGB at 40, 30, and 20 dB SNR while the LR-HSI was left clean, and the SSF was retrained from scratch at each level; every condition was run with three random seeds under an identical 100k-epoch, constant-1e-4 protocol. Table VIII reports the reconstruction quality against the clean baseline. Degradation is graceful across the milder settings—SAM rises only from 2.31 on the clean input to 2.55 at 40 dB and 3.75 at 30 dB, with PSNR and ERGAS almost unchanged—and becomes pronounced only at the heaviest 20 dB corruption, where SAM reaches 7.30 and ERGAS climbs to 16.14. The very small inter-seed variance (SAM standard deviation below 0.02 at every level) indicates that this behaviour is a stable property of the method rather than an artefact of initialisation or a particular noise draw. The method therefore remains usable under the moderate noise typical of a consumer camera and exhibits a predictable, monotonic decline as the input is degraded further.

**TABLE VIII**

*UISSF reconstruction on Chikusei under increasing HR-RGB noise (LR-HSI kept clean; mean ± std over three seeds)*

| SNR | SAM↓ | PSNR↑ | ERGAS↓ |
|---|---|---|---|
| Clean | 2.31 ± 0.02 | 52.87 ± 0.07 | 6.79 ± 0.00 |
| 40 dB | 2.55 ± 0.01 | 51.70 ± 0.04 | 6.99 ± 0.00 |
| 30 dB | 3.75 ± 0.01 | 47.89 ± 0.00 | 8.50 ± 0.02 |
| 20 dB | 7.30 ± 0.01 | 41.57 ± 0.01 | 16.14 ± 0.04 |

*Clean denotes the noise-free HR-RGB. The noise study is conducted on Chikusei, the airborne remote-sensing scene most representative of the low-cost-sensor setting motivating the analysis.*

## VI. Conclusion

We presented a correspondence-free principle for spectral mapping and used it to perform hyperspectral image super-resolution directly from an HR-RGB, independent of both the spatial registration and the CRF relating the inputs. The principle rests on a single observation: the abundance Gram matrix is invariant to pixel permutation and therefore supplies a supervisory signal that constrains spectral composition without spatial correspondence. Because the operator is indifferent to mapping direction, it unifies the well-posed contraction from spectra to RGB, established earlier as CRF learning [23], with the ill-posed expansion from RGB to spectra solved here; the two directions are instances of one principle rather than separate methods. Across indoor, natural-scene, and remote-sensing benchmarks, the unsupervised and sensor-independent (CRF-free) SSF matched state-of-the-art methods while requiring neither of the assumptions those methods depend on, and it remained accurate when learned from fully pixel-permuted HR-RGB after inverse reindexing for evaluation. Inference consists of a single feed-forward pass through the learned SSF and the frozen decomposition network. The remaining limitation is accuracy on spectrally atypical, low-signal content—most visibly water bodies and, more broadly, the CAVE benchmark—where the shared endmember representation is less expressive; this is the natural cost of a correspondence-free, CRF-free formulation, and unlike dependence on registration or a known CRF it does not compromise deployability. The loss ablation reinforced the central claim: single- and multi-term objectives reached comparable accuracy on Chikusei, indicating that the method's effectiveness derives from the permutation-invariant Gram-matrix supervision itself rather than from a carefully weighted loss. The magnitude of the reconstruction is fixed primarily by the architecture—the sum-to-one abundance constraint and the physically scaled endmembers—leaving the loss to constrain spectral structure. Future work will investigate extending the proposed correspondence-free supervision beyond hyperspectral super-resolution to other multimodal reconstruction tasks where reliable cross-modal correspondence cannot be assumed.

## Acknowledgment

This study was partially supported by the Ministry of Agriculture, Forestry and Fisheries (MAFF) "Research project for technologies to strengthen the international competitiveness of Japan's agriculture and food industry" and the Japan Science and Technology Agency (JST) AIP Acceleration Research (JPMJCR21U3). This work was also supported by the Norwegian Space Agency, grant number 74CO2210 ("Bruk av Copernicus/Sentinel-data til jordbruksfaglige problemstillinger").